\documentclass{isprs} 
\usepackage{subfigure}
\usepackage{setspace}
\usepackage{geometry} 
\usepackage{epstopdf}
\usepackage[labelsep=period]{caption}  
\usepackage[british]{babel} 
\usepackage[hang]{footmisc}
\usepackage{amsmath}
\usepackage{multirow}
\usepackage{pifont}
\usepackage{makecell}
\usepackage[authoryear]{natbib}

\begin{document}

\linespread{1.08333}

\title{Depth-Aware Panoptic Segmentation}
\date{}

\author{T. Nguyen\thanks{Corresponding author} , M. Mehltretter, F. Rottensteiner}

\address{**** (for review, affiliations must be rendered anonymous)}
\address{Institute of Photogrammetry and GeoInformation, Leibniz University Hannover, Germany\\ 
         (tuan.nguyen, mehltretter, rottensteiner)@ipi.uni-hannover.de}


\commission{II, }{2} 
\workinggroup{II/3} 
\icwg{}   

\abstract{
Panoptic segmentation unifies semantic and instance segmentation and thus delivers a semantic class label and, for so-called {\it thing} classes, also an instance label per pixel.
The differentiation of distinct objects of the same class with a similar appearance is particularly challenging and frequently causes such objects to be incorrectly assigned to a single instance.
In the present work, we demonstrate that information on the 3D geometry of the observed scene can be used to mitigate this issue:
We present a novel CNN-based method for panoptic segmentation which processes RGB images and depth maps given as input in separate network branches and fuses the resulting feature maps in a late fusion manner.
Moreover, we propose a new depth-aware dice loss term which penalises the assignment of pixels to the same {\it thing} instance based on the difference between their associated distances to the camera.
Experiments carried out on the Cityscapes dataset show that the proposed method reduces the number of objects that are erroneously merged into one {\it thing} instance and outperforms the method used as basis by $+2.2\%$ in terms of panoptic quality.
}


\maketitle



\section{Introduction}
\label{INTRODUCTION}

\sloppy

Panoptic segmentation combines the tasks of semantic segmentation and instance segmentation \citep{kirillov2019panoptic}. 
For a set of {\em thing} classes,  e.g.  {\em car}, it delivers information about individual instances, e.g.\ in the form of bounding boxes with class labels and binary masks indicating the pixels corresponding to the instance. 
Image regions not belonging to {\em thing} instances ({\em background} in instance segmentation)  are assigned to one of the so-called {\em stuff} classes in a similar way as in semantic segmentation. 
For these classes (e.g.,  {\em wall}), no information about instances is determined. 

\begin{figure}[ht!]
\begin{center}
		\includegraphics[width=1.0\columnwidth]{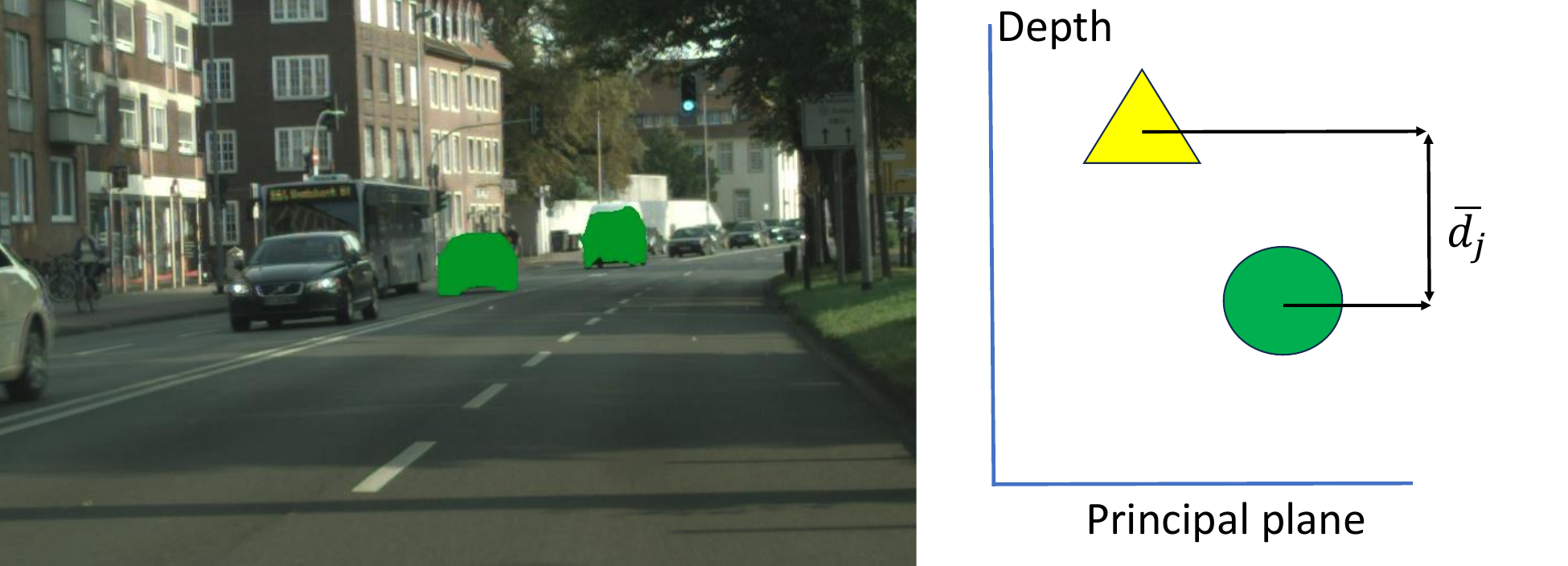}
	\caption{\textit{Left:} A binary instance mask predicted by \protect\citep{li2021fully} which erroneously merges two  {\em car} instances, superimposed to the input. \textit{Right:} We exploit the depth difference $\bar{d_j}$ between pixels corresponding to different instances (the \textit{triangle} and the \textit{circle}) in training to mitigate the problem.}
\label{fig:ambiguity_case_and_our_idea}
\end{center}
\end{figure}

This task is usually solved using neural networks. 
Early approaches merged the results of separate methods for instance and semantic segmentation in  post-processing  \citep{kirillov2019panoptic}. 
Recent approaches apply unified strategies that  allow for end-to-end training. 
\citet{li2021fully} achieve  this goal by predicting a binary mask for every {\em stuff} class and a binary mask and a class label for every instance of every {\em thing} class. 
This  avoids the need for bounding box proposals and  allows the  network to learn the two sub-tasks jointly in an end-to-end manner.  
Existing work  usually only relies on RGB images as input. Fig.~\ref{fig:ambiguity_case_and_our_idea} shows an example for a binary instance mask predicted from such an image by  \citep{li2021fully}. 
In this example, two {\em car} instances having a similar appearance are actually merged into one. 
One way to overcome such problems is to integrate additional information. 
In this work, we  utilise {\em depth} from stereo images as an additional input, thus proposing a method for depth-aware panoptic segmentation. 

RGB and depth data have been used for semantic and instance segmentation for some time, but there are only few works on panoptic segmentation exploiting both modalities. 
\citet{narita2019panopticfusion} still apply RGB images for panoptic segmentation, using depth only to estimate camera poses and to generate a 3D map. 
\citet{seichter2022efficient} process depth and colour in two separate branches of a deep neural network before fusing the outputs to use them  for instance segmentation and semantic segmentation.
Thus, this method still involves two separate networks. 
In this work, we try to overcome some of the problems of existing methods, making the following scientific contributions: 
\begin{itemize}
    \item We propose a method for the joint use of colour and depth for panoptic segmentation that can be trained  end-to-end.
    \item In this context, we investigate two different techniques for the fusion of the colour and depth branches of the network. 
    \item We propose a new depth-aware loss  term in order to mitigate problems such as the one indicated in Fig.~\ref{fig:ambiguity_case_and_our_idea}, exploiting the depth differences of separate {\em thing} instances. 
    \item We show the improvements achieved by the additional information and the new loss function in experiments using a publicly available benchmark dataset. 
\end{itemize}

Our method is based on \citep{li2021fully}, which we extend by an additional depth branch, fusing the resultant features with those obtained from the colour branch of the network, and by a new depth-aware loss function. 


\section{Related Work}\label{sec:RELATED WORK}

Panoptic segmentation methods can be divided into top-down (box-based), bottom-up (box-free) and unified approaches, the latter being also referred to as single path approaches.
Top-down and bottom-up approaches treat  semantic and instance segmentation separately before merging their results to obtain the panoptic segmentation results. 
Top-down methods follow a two-stage design and estimate bounding boxes for {\em thing} instances first, before a pixel-wise mask and a semantic label is predicted per instance, e.g.\ \citep{kirillov2019panopticfpn, xiong2019upsnet, li2019attention}. 
The semantic segmentation of the background (i.e.\ all pixels not corresponding to a {\em thing} instance) is commonly carried out separately.
As a consequence, the performance highly depends on the quality of the estimated bounding boxes. 
Consistency between the semantic segmentation masks of overlapping bounding boxes and between instances and the background is not guaranteed, which requires to resolve conflicts in a heuristic post-processing step.
Bottom-up approaches address this limitation by estimating semantic and instance segmentation masks without relying on previously estimated bounding boxes. 
For instance, \citet{cheng2020panoptic} apply semantic segmentation differentiating both  {\em stuff} and  {\em thing} classes. 
The instance masks are derived from the outputs of two additional network branches: a centre point for each object instance and an offset to the corresponding centre point for each pixel being located on such an instance. 
All network branches can be trained end-to-end, but some rather complex post-processing is required to derive the instance masks and class labels from the original output. 

Unified approaches do not apply separate networks or network branches for semantic and instance segmentation, but solve the panoptic segmentation task  directly, e.g. by simultaneously predicting binary masks for {\em stuff} classes and {\em thing} instances.  
Following this strategy, \citet{li2021fully} propose to learn the estimation of two types of intermediate feature maps: 
maps that describe individual \textit{thing} instances and \textit{stuff} classes and maps that encode the input image.
Maps of the first type are used to extract filter kernels for convolutions that are applied to the second type of  maps. 
The result of these convolutions is a set of binary  masks (one  per \textit{thing} instance  and one  per \textit{stuff} class).
A limitation of \citep{li2021fully} is the incorrect assignment of pixels showing distinct objects of similar appearance to a single instance mask (c.f.~Fig.\,\ref{fig:ambiguity_case_and_our_idea}).
\citet{de2023intra} claim  that this problem is related to the   training procedure which only uses image crops, as only a small number of (partially visible) objects is seen by the network at once.
They propose an additional loss term that enforces the two kinds of feature maps described above to be different for each image crop, assuming that different crops show different objects.
\citet{zhang2021k} propose a method similar to \citep{li2021fully}, estimating and using the previously mentioned two types of features maps in the same way. Building on \citep{kirillov2019panopticfpn}, the authors focus on the discriminative ability of the feature maps used as filter kernels by following a clustering-based approach, which encourages features from the same class to be similar and features from different classes to be distinct.
\citet{wang2021max} propose an attention-based architecture with a 2D pixel-based and a 1D global memory path. 
The former is used to estimate a binary segmentation mask per instance, the latter provides a semantic class label per mask. The two paths are densely connected with so-called dual-path transformer blocks, which allow to interchange information between the two paths. 
To ensure consistency across the individual segmentation masks, i.e., that each pixel of an image belongs to exactly one mask, the softmax function is applied per pixel to the set of predicted segmentation masks.
\citet{yu2022cmt} incorporate the concept of conventional clustering approaches into a mask transformer architecture to identify pixels that belong to the same object instance in an early stage of the neural network. 
The assignment of pixels to clusters as well as the update of cluster centres and per pixel feature descriptors are realised as attention layers and are  computed iteratively.
Unified approaches could achieve a significant improvement in panoptic segmentation, also reducing the need for post-processing to obtain a consistent result. 
However, the majority of methods rely on a single RGB image, being thus limited to 2D information on the observed scene.

To improve the results even further, information about the 3D geometry of the observed scene can used as an additional input.  
\citet{narita2019panopticfusion} take a sequence of RGB images and corresponding depth maps as input to estimate a panoptic segmentation in 3D in form of a volumetric map. 
Panoptic segmentation is first carried out in 2D per frame of the sequence, using one RGB image only. The depth information is used to estimate the exterior orientation parameters of the corresponding RGB image and to combine the frame-based 2D panoptic segmentation masks into a volumetric 3D representation for the whole sequence.
\citet{wu2021scenegraphfusion} present a method for incremental 3D scene graph estimation from RGB and depth data that also delivers a panoptic segmentation of the observed 3D surfaces as a by-product. 
A graph neural network is trained  to build a graph in which clusters of pixels that belong to the same object or object part correspond to the nodes, while the edges represent  geometric relations between the nodes.
A panoptic segmentation is obtained by combining nodes corresponding to the same object based on the edge information.
In both of these methods,  depth  is not used to support the panoptic segmentation itself, but only to fuse independently estimated 2D panoptic segmentation masks and to lift these masks from 2D to 3D.
In contrast, \citet{seichter2022efficient} use both, an RGB image and a depth map, to perform a 2D panoptic segmentation using an encoder-decoder architecture. 
First, colour and depth are processed in two separate encoder branches.
The extracted feature maps are fused at different scales, making this an example for a late fusion approach. 
The decoder consists of two separate branches as well, one for estimating a semantic and one for an instance segmentation. 
Thus, \citet{seichter2022efficient} follow a bottom-up strategy, which  suffers from the limitations discussed earlier. 
While depth information is used for panoptic segmentation, the relation between depth and the segmentation masks to be estimated is learned in a purely data-driven way, i.e., no constraints on the segmentation are explicitly introduced based on the geometry in training.

In summary, \citep{li2021fully} and \citep{seichter2022efficient} can be considered the most similar works to the one  presented in this paper. 
We use \citep{li2021fully} as the basis for our work, but extend it by incorporating an additional branch for processing depth information and by the loss function used in training, for which we propose a new depth-aware term. 
The way in which we integrate depth is inspired by \citet{seichter2022efficient}, but our overall architecture is different. 
Furthermore,  depth is not just used as as additional input, but also in the loss function to explicitly constraint the assignment of pixels to instance masks.



\section{Background: Panoptic FCN} 
\label{sec:panoptic_fcn}
\begin{figure*}[ht]
\begin{center}
		\includegraphics[width=\textwidth]{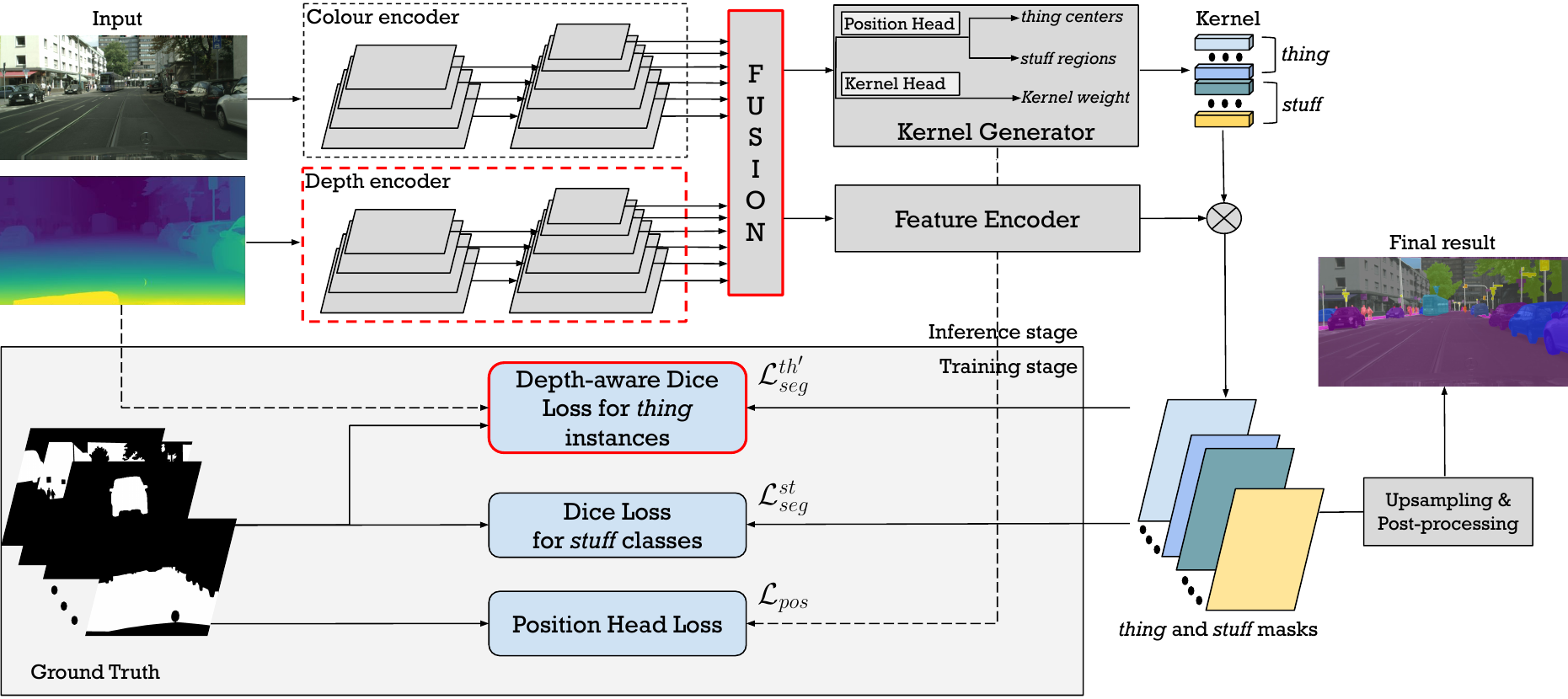}
	\caption{Our proposed method. The blocks with a red edging are our proposed modules. The remaining ones are  also used in Panoptic FCN, but there the output of the colour encoder is directly processed by the feature encoder and kernel generator blocks \protect\citep{li2021fully}. 
    Our method additionally uses an encoder for the depth map and a fusion module; the subsequent blocks process the results of colour and depth fusion.  $\otimes$ indicates a convolution. 
    In training, we use a new depth-aware Dice loss for the {\em thing} instances.}
\label{fig:architecture}
\end{center}
\end{figure*}

To make this paper self-contained, we start with a brief summary of Panoptic FCN \citep{li2021fully}. 
It uses RGB colour images $X^{c}\in R^{3\times H \times W}$ as input, where $H$ and $W$ represent the image height and width,  respectively. 
The goal is to assign every pixel of a picture either to  one of $K^{st}$ {\em stuff} classes or to an instance of one of $N^{th}$ {\em thing} classes. 
Every image is presented to a Feature Pyramid Network (FPN) with Resnet50  backbone \citep{lin2017feature} to extract features at different scales, resulting in a set of feature maps $P_p$, $p\in\{2, \ldots, 7\}$, with spatial extents $H_p \times  W_p = H/(2^p) \times  W/(2^p)$. 
The outputs of the FPN  are processed further in two separate branches: the {\em Feature Encoder} and the  {\em Kernel Generator}, cf.\  Fig.~\ref{fig:architecture}  \citep{li2021fully}. 

In the Feature Encoder, the feature maps $P_2$ to $P_5$ are first processed by  the semantic FPN module of \citet{kirillov2019panopticfpn} and then by three sequential convolution layers. 
The result of the last layer is a feature map  of dimension ${C_e\times H/4 \times W/4}$ encoding the image content in a way appropriate for the task. 

The  input of the Kernel Generator consists of the feature maps $P_3$ to $P_7$ generated by the FPN. 
First, each feature map $P_p$ is processed independently by two  heads, each consisting of three sequential convolution layers \citep{li2021fully}: the  {\em Kernel Head} and the {\em Position Head}.
The {\em Kernel Head} is trained to predict the {\em kernel weight} tensor of dimension  ${C_e\times H/4 \times W/4}$, which contains a  weight vector (referred to as {\em kernel}) for every spatial position of the feature map $P_p$. 
The output of the last convolution layer of the {\em Position Head} consists of ($N^{th}$ + $K^{st}$) maps of class scores, i.e.\ one per class (normalised by a sigmoid function). 
For the $K^{st}$ {\em stuff} classes, each  map contains the probability for every pixel in $P_p$ to belong to the corresponding class; applying  a threshold, that map is converted into a binary map indicating the pixels of that class. 
For the $N^{th}$ {\em thing} classes, these maps indicate the probability of a pixel to correspond to a centre of a  {\em thing} instance. 
Instance centres are determined by  applying a threshold and local nonmax suppression to these maps \citep{zhou2019objects}. 

The output of the Position Head is used to define the kernels that are the output of the Kernel Generator. 
For every {\em stuff} class, at every scale $p$, one kernel is obtained by computing the average of the vectors in the kernel weight tensor at the positions assigned to that class in the binary map generated by the Position Head, and the kernels determined at different scales are averaged. 
For the {\em thing}  classes, the kernel related to an instance at a scale $p$ is extracted from the kernel weight tensor at the position of the instance  centre. 
The resultant {\em thing} kernels extracted at different scales have to be combined.
To do so, kernels related to instances of the same  {\em thing} class are merged by averaging if their cosine similarity is above a threshold. 
This will lead to $K^{th}_0$ {\em thing} instances for which the kernel and the class label are known. 
If $K^{th}_0$ is larger than a pre-defined value $K^{th}_{max}$ (set to 100 in the experiments), the kernels are ordered according to the confidence scores from the heatmap, and the $K^{th}_{max}$ kernels having the highest confidence are preserved, thus $K^{th}=K^{th}_{max}$. 
Otherwise, $K^{th}$ is set to $K^{th}=K^{th}_{0}$. 
The final output of  the Kernel Generator consists of the $K^{st}+K^{th}$ kernels of dimension $1 \times C_e$, each associated with  a ({\em stuff} or {\em thing}) class label. 

The output of the Feature Encoder is convolved with each of the  kernels, and each of the outputs is normalised by a sigmoid function, yielding $K^{st}+K^{th}$ maps of class scores for pixels to belong to one of the {\em stuff} classes or to one of the {\em thing} instances at a reduced resolution ($H/4\times W/4$).
The class label associated with a kernel is also associated with the corresponding mask. 
These masks are upsampled by bilinear interpolation to obtain scores at the original resolution, and after applying a threshold, $K^{st}+K^{th}$ binary masks are generated that indicate whether a pixel belongs to the corresponding {\em stuff} class or to the corresponding instance of a {\em thing} class. 

Finally, post-processing  is applied to remove contradictions between the predicted binary maps in a way similar to \citep{kirillov2019panopticfpn}. 
Pixels  not assigned to any of the classes or instances are considered to be {\em background}. 
The resultant $K^{st}+K^{th}$ binary maps of size $H \times W$ along with the corresponding class labels are the final output.

For training, a reference consisting of binary masks for the {\em stuff} classes and the {\em thing} instances is required. 
Training is based on minimising a loss function $\mathcal{L}$   \citep{li2021fully}: 
\begin{equation}\label{equ:0}
	\mathcal{L} = \lambda_{pos} \cdot\mathcal{L}_{pos} + \lambda_{seg} \cdot\mathcal{L}_{seg}, 
\end{equation}
where $\lambda_{pos}$ and $\lambda_{seg}$ are hyperparameters for weighting the two loss terms.
The  term $\mathcal{L}_{pos}$ is applied to the output of the Position Head of the network. 
It compares the maps containing class scores determined for every scale to a reference using a focal loss \citep{lin2017focal}. 
The reference for the {\em stuff} classes consists of the binary masks downsampled by a factor of 4. 
In case of the {\em thing} instances, for every class a binary mask showing the centres of all reference instances of that class is generated first. 
The reference for the {\em thing} centres is obtained by blurring this mask. 
Consequently, this reference is not binary.

The second loss term, $\mathcal{L}_{seg}$, is applied to the sigmoid scores predicted at the resolution $H/4 \times W/4$, i.e. before upsampling. 
Consequently, the reference maps have to be downsampled by a factor of 4 for training. 
In training, the kernels are not determined on the basis of the predictions of the Position Head, but they are sampled on the basis of the reference. 
For the {\em stuff} classes, one position is randomly sampled inside of the area assigned to that class in the reference at the corresponding scale, and the kernel related to that scale is sampled at that random position. 
For {\em thing} instances, the $k$ pixels inside the instance according to the reference mask having the highest confidence in the prediction are used to extract the kernel at every scale (\citet{li2021fully} use $k=7$). 
In this way, it is known which  predicted  instance masks correspond to which reference  masks. 
$\mathcal{L}_{seg}$ is modelled as a Dice loss  \citep{milletari2016v}, comparing the binary masks for all {\em stuff} classes and  {\em thing} instances. 


\section{Depth-aware Panoptic Segmentation}
\label{sec:METHODOLOGY}

We start the presentation of our  method for depth-aware panoptic segmentation with an overview (Section \ref{subsec:overview}). 
Afterwards, we focus on our main modifications compared to the baseline (cf.\ Section~\ref{sec:panoptic_fcn}):  
our concept of fusing RGB and depth data is presented in Section \ref{subsec:colordepthfusion}, while  the training procedure, introducing our new depth-aware dice loss, is described in Section~\ref{subsec:depthawarediceloss}.


\subsection{Overview} 
\label{subsec:overview}

Our method is based on Panoptic FCN as presented in Section~\ref{sec:panoptic_fcn}, expanding it so that it can use a depth map as an additional input. 
The architecture is shown in Fig.~\ref{fig:architecture}, which also highlights our new contributions by red edging. 
The input consists of a colour (RGB) image  $X^{c} \in R^{3\times H \times W}$ and a corresponding depth map $X^{d} \in R^{1 \times H \times W}$ of the same size and given in the same coordinate frame. 
In principle, any method can be used to generate the depth map; we used stereo matching in our experiments. 
We decided to use a late fusion approach in which the colour and depth images are processed in separate encoder branches before being fused in order to generate the feature map that serves as the input to the Feature Encoder and to the Kernel Generator. 
Details about this fusion approach are presented in Section~\ref{subsec:colordepthfusion}. 
For the Feature Encoder and the Kernel Generator we use the architecture  described in  Section~\ref{sec:panoptic_fcn}.
The output of our method also consists of $K^{st}$ binary maps identifying all pixels of the {\em stuff} classes and $K^{th}$ binary maps identifying all pixels corresponding to one of the instances of the {\em thing} classes, in the latter case along with the class labels. 

Training is also based on minimising the loss function having two components according to equ.~\ref{equ:0}. 
However, in order to alleviate problems such as those indicated in Fig.~\ref{fig:ambiguity_case_and_our_idea}, we propose a new depth-aware Dice loss that is applied to the \textit{thing} instances in our model for the loss term $\mathcal{L}_{seg}$.
The training procedure and this new loss function are explained in Section~\ref{subsec:depthawarediceloss}.



\subsection{Colour and Depth Fusion} \label{subsec:colordepthfusion}

\citet{seichter2021efficient} process the colour and  depth images in  separate encoder branches with a similar architecture before fusing the resultant features. 
We follow this {\em late fusion approach}, extending the Panoptic FCN architecture by a depth branch in the encoder. 
We do so because in preliminary experiments this variant outperformed an early fusion approach in which the depth map was just concatenated to the RGB image presented to the FPN backbone as a fourth input band. 
The depth branch has the same architecture as the colour branch, except that the input only consists of a single band. 
Thus, the two encoder branches deliver two  multi-scale outputs, $P^c_p$ and $P^d_p$  for colour and depth, respectively, with $p\in\{2, \ldots, 7\}$ and dimensions as described for the colour backbone in Section~\ref{sec:panoptic_fcn}. 

The fusion block in Fig.~\ref{fig:architecture} combines the colour and depth feature maps at corresponding scales to obtain fused feature maps $P^f_p$.
There are several ways in which the fusion can be carried out. 
Our default option is  \textit{mean} fusion: 
\begin{equation}\label{equ:1_1}
	P^f_p = (P^c_p + P^d_p) / 2~~\forall~p \in \{2, \ldots, 7\}. 
\end{equation}
In this case, the fused feature map at scale level $p$ is determined as the arithmetic mean of the corresponding colour and depth feature maps. 
In our experiments, we compare this default method to fusion based on  \textit{concatenation}: 
\begin{equation}\label{equ:1_2}
	P^f_p = conv \left (concatenate(P^c_p, P^d_p) \right)~\forall~p \in \{2, \ldots, 7\}
\end{equation}
Here, two colour and depth feature maps at scale level $p$ are concatenated first. 
After that, a point-wise ($1 \times 1$) convolution ($conv$) is applied to reduce the number of features to $C_e$, i.e. the number of features of each of the input maps (cf.\ Section~\ref{sec:panoptic_fcn}).  

In preliminary experiments, similarly to \citet{seichter2021efficient}, we also tested  fusion based on Squeeze-and-Excitation blocks \citep{hu2018squeeze}. 
However, while requiring more  parameters, it did not give better results than  \textit{mean} and \textit{concatenation}  fusion, so that it is not considered in this paper. 


\subsection{Training and Depth-aware Dice Loss} \label{subsec:depthawarediceloss}

As in the baseline, the loss minimised in training consists of two terms (cf.\ Section~\ref{sec:panoptic_fcn}, equ.~\ref{equ:0}).
The component $\mathcal{L}_{pos}$ used to constrain the output of the Position Head is identical to the one used in \citep{li2021fully}. 
However, we modify the term  $\mathcal{L}_{seg}$, i.e. the loss applied to the output of the panoptic segmentation.
\citet{li2021fully} use a loss based on the Dice function \citep{milletari2016v} which measures the level of agreement of two binary  images $Pr$ and $Gt$ of equal size: 
\begin{equation}\label{equ:3}
    \begin{split}
        Dice(Pr, Gt) = \frac{2\cdot\sum^N_{j=1} p_j\cdot g_j}{\sum^N_{j=1} p^2_j + \sum^N_j g^2_j}
    \end{split},
\end{equation}
where $p_j \in\{0,1\}$ is the grey value of the $j^{th}$ pixel in the predicted mask  $Pr$,  $g_j\in\{0,1\}$ is the corresponding grey value in the ground truth mask $Gt$, and $N$ is the number of pixels in the masks.
As the Dice function according to equ.~\ref{equ:3} measures similarity, 
the Dice loss is based on $1 -Dice(Pr, Gt)$. 

However, Panoptic FCN trained using the Dice loss for $\mathcal{L}_{seg}$  occasionally delivers instance masks that contain two spatially separated {\em thing} objects of the same type if the latter have a similar appearance (e.g.\ Fig.~\ref{fig:ambiguity_case_and_our_idea}). 
To address this problem, we introduce a new term into the loss $\mathcal{L}_{seg}$ which utilises depth information to penalise the assignment of a pixel  to a   {\em thing}  instance if the absolute difference between its depth value and the average depth of the instance according to its extents in the ground truth is large. 
In  this way, the network can learn that pixels within one instance of a {\em thing}  class  have  similar depth values. 
In the original Dice Loss, a false positive (FP) pixel $p_j$  in the prediction mask (indicated by $g_j =0$ and  $p_j=1$) will decrease the output of the Dice function (equ.~\ref{equ:3}), because that pixel will increase the denominator by 1 while not increasing the numerator. 
Thus, a FP pixel will increase the loss. 
Our idea is to increase the loss even further for FP pixels that are at a depth  different from the one of the instance. 
This can be achieved by a loss based on a new depth-aware Dice function {\em DDice} defined as:
\begin{equation}\label{equ:4}
    \begin{split}
        DDice(Pr, Gt, d) = \frac{2\cdot\sum^N_{j=1} p_j \cdot g_j  }{\sum^N_{j=1} \left[p_j\cdot\left(1+\omega\cdot\Bar{d}_j\right)\right]^2 + \sum^N_j g^2_j},
    \end{split}
\end{equation}
where $Pr$ and $Gt$ are a predicted and a ground truth binary map for a specific {\em thing} instance, $p_j$ as well as $g_j$ are the corresponding grey values at pixel $j$, $N$ is the number of pixels in a map, and $d$ is a depth map having the same size as $Pr$ and $Gt$. 
The desired depth-awareness is achieved by the factor $\left(1+\omega\cdot\Bar{d}_j\right)$ in the denominator. 
Here, $\omega$ is a hyperparameter  modulating the impact of the depth on the loss and $\Bar{d_j}$ is based on the difference of the   depth $d_j$ of a FP pixel $j$  from the mean depth $d_g$ of  the pixels assigned to the instance corresponding to $Gt$: 
\begin{equation}\label{equ:dj}
\Bar{d}_j=\left|\frac{d_j-d_g}{\max \left(d_g, d_{max}-d_g\right)}\right| \cdot p_j \cdot\left(1-g_j\right), 
\end{equation}
with
\begin{equation}\label{equ:dg}
d_g = \frac{1}{\sum^N_{j=1} g_j} \cdot \sum^N_{j=1} g_j\cdot d_j. \nonumber
\end{equation}
In equ.~\ref{equ:dj}, $d_{max}$ denotes a hyperparameter corresponding to the maximum possible depth value. 
As the product $p_j\cdot\left(1-g_j\right)$ is 0 except for FP pixels,  the depth-dependent term $\omega\cdot\Bar{d}_j$ only decreases the output (and, thus, increases the loss) for FP pixels. 
Note that for $\omega = 0$, our depth-aware Dice function is equivalent to the Dice function in equ~\ref{equ:3}. 

In the training stage, after estimating $K^{st}$ masks for {\em stuff} classes and  $K^{th}$ {\em thing} instance masks, the loss term $\mathcal{L}_{seg}$ is evaluated and used to update the network parameters. 
For the {\em stuff} classes, we use the standard Dice loss based on equ.~\ref{equ:3} to define a term $\mathcal{L}_{seg}^{st}$,  whereas for {\em thing} instances, a loss $\mathcal{L}_{seg}^{th{'}}$ based on our depth-aware dice loss (equ.~\ref{equ:4}) is applied. 
We obtain the following formulation for the loss  $\mathcal{L}_{seg}$ in equ.~\ref{equ:0}:
\begin{equation}\label{equ:5}
    \begin{split}
        \mathcal{L}_{seg} &= \mathcal{L}_{seg}^{st} + \mathcal{L}_{seg}^{th{'}} = \\
                & =\frac{1}{K^{st}}\cdot\sum_{k_{st}=1}^{K^{st}}\left[1-Dice\left(Pr_{k_{st}}, Gt_{k_{st}}\right)\right] \\
                & + \frac{1}{K^{th}}\cdot\sum_{k_{th}=1}^{K^{th}}\left[1-DDice\left(Pr_{k_{th}}, Gt_{k_{th}}, X^{d}\right)\right].  
    \end{split}
\end{equation}
This is different from \citet{li2021fully}, who also use the $Dice$ function to model  the loss component for the {\em thing} instances.  
Equ.~\ref{equ:5} gives the loss term $\mathcal{L}_{seg}$ for a single training image with corresponding depth map $X^d$. 
$Pr_{k_{st}}$ and $Gt_{k_{st}}$ are the predicted and ground truth  maps for the {\em stuff} class $k_{st}$ for that training image. 
Similarly, $Pr_{k_{th}}$ and $Gt_{k_{th}}$ are the predicted and ground truth  maps for the $k_{th}$  {\em thing} instance. 
Note that, due to the way in which the instance centres are initiated at training time (cf.\ Section~\ref{sec:panoptic_fcn}), for each predicted map it is known to which reference instance it corresponds.
There is no need for matching instance predictions to ground truth instance maps to establish which predicted instance map is considered to correspond to the ground truth instance $k_{th}$. 
The actual loss used in training is a sum over all images of a minibatch. 

\begin{figure}[ht]
\begin{center}
		\includegraphics[width=1.0\columnwidth]{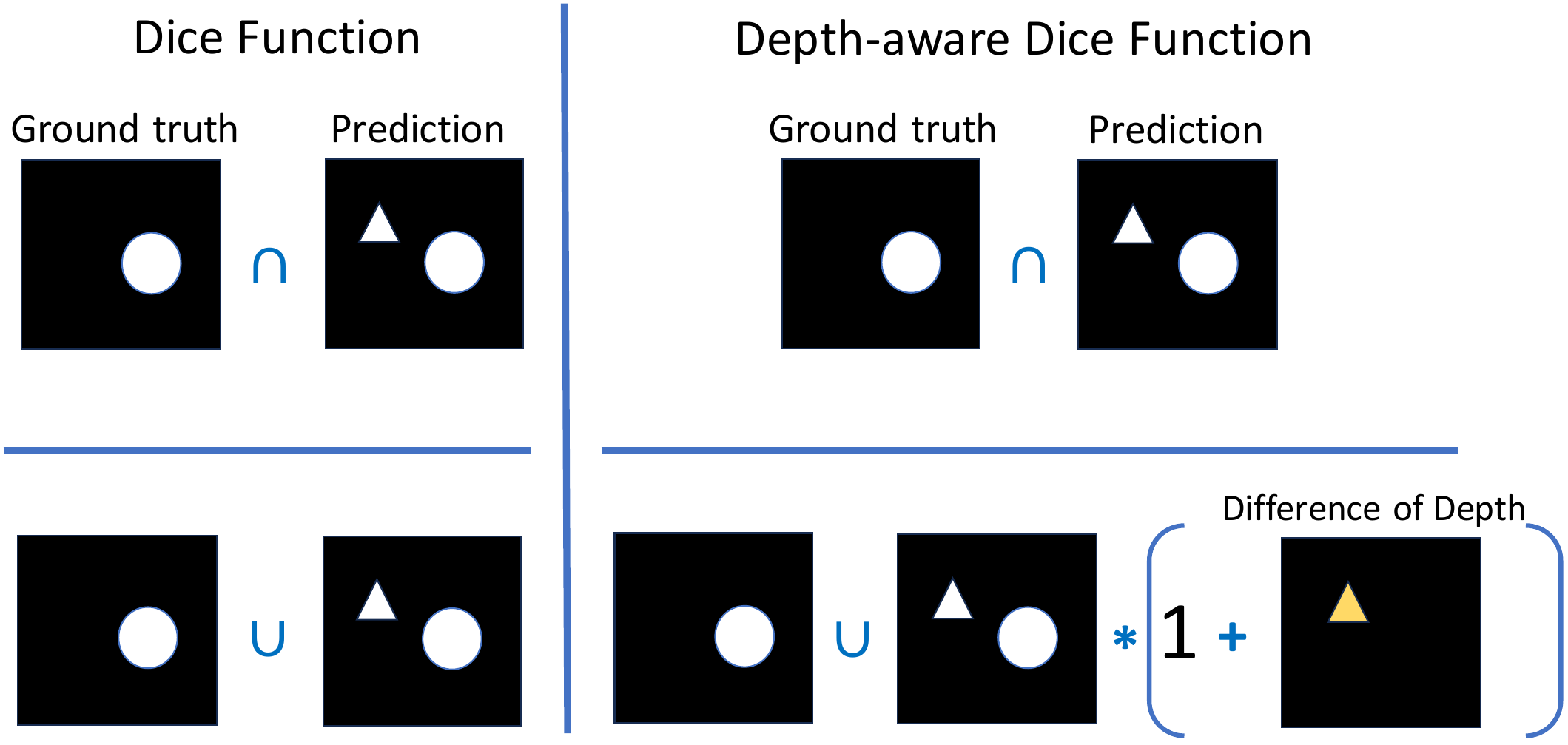}
	\caption{Visualisation of the  Dice function (left) and the new depth-aware variant (right). In the latter case, the consideration of  depth  will increase  the penalty  for FPs with large depth differences compared to the TPs.}
\label{fig:depth_aware_dice_loss}
\end{center}
\end{figure}

The Dice loss and our  depth-aware Dice loss are visualised in Fig.~\ref{fig:depth_aware_dice_loss}, where the circle   represents the true positive (TP) pixels  of an instance and the triangle corresponds to FPs. 
Our new loss function penalises  FP pixels having a large depth difference from the mean of the TPs. 
The larger the difference in depth between the FP segments and the ground truth, the larger the penalty that is added for this segment. 

\section{Experiments}\label{sec:EXPERIMENTS}

We start the presentation of our experiments by  introducing the experimental setup in Section~\ref{subsec:Setup}.
The results achieved by our method are described in Section~\ref{subsec:evaldepthawarediceloss}, while  
Section~\ref{subsec:Ablation} presents two ablation studies. 


\subsection{Experimental Setup} \label{subsec:Setup}


\subsubsection{Dataset: } \label{subsubsec:Dataset}
We perform our experiments on the Cityscapes dataset \citep{cordts2016cityscapes}. 
It consists of $5$k stereo image pairs showing various street scenes in  Germany. 
The size of all images is $1024 \times 2048$ pixels, and panoptic labels are provided for the left image of every stereo pair, considering  $K^{st}=11$ \textit{stuff} and $N^{th}=8$ \textit{thing} classes. 
The dataset is split into training, validation and test sets. 
We used the training set consisting of  $2975$ images for training our method. 
As the reference is unavailable for the test set, we followed the experimental protocol of our baseline methods \citep{lipson2021raft,de2023intra} and used the validation set, consisting of $500$ image pairs, for testing. 

The Cityscapes dataset also provides disparity maps for every stereo pair, computed with a variant of Semi-global Matching (SGM) \citep{cordts2016cityscapes,hirschmuller2007stereo}, from which we derived depth maps.
However, we found the  SGM-based depth maps to contain a considerable number of incorrect depth estimates and  relatively large regions without any meaningful depth values. 
Thus, the relevance of these depth maps for the classification was expected to be small, which was confirmed in preliminary experiments. 
We decided to generate better depth maps, using RAFT-stereo \citep{lipson2021raft}.
The resulting depth maps, containing depth values in $[m]$, were post-processed by filtering out unreasonably small or large depth values, setting depth values smaller than $d_{min}=1\,m$ and larger than $d_{max}=500\,m$  to $0$. 



\subsubsection{Experimental Protocol: } \label{subsubsec:experimentalsetup} 

Training is based on minimising the loss according to equs.~\ref{equ:0} and~\ref{equ:5}. 
Similarly to \citet{li2021fully}, we use Stochastic Gradient Descent with a weight decay of $10^{-4}$ and a momentum of $0.9$ for that purpose. 
We also follow the baseline method by applying data augmentation, using minibatches consisting of patches of $512 \times 1024 $ pixels. 
These patches are randomly cropped from the input after scaling the images and depth maps  by a random factor  $f\in[0.5, 2]$ and applying a random horizontal flip. 
The depth values are also scaled to maintain the ratio between the extents in the image plane and the depth. 
The input patches are normalised by subtracting the channel-wise means $\mu$ and dividing the differences by the channel-wise standard deviations $\sigma$, these values being computed from all images and depth maps in the training set, respectively. 
Note that pixels marked as not being in the depth range $[d_{min}, d_{max}]$ are not considered to determine $\mu$ and $\sigma$.
A minibatch  that is processed in one training iteration consists of $12$ such patches,  and $180$k such iterations are carried out.
The learning rate is initially set to $0.02$ and reduced by a factor of $0.9$ after every 1000$^{th}$ iteration. 
The parameters of the colour and depth encoders  are both initialised by values obtained from pretraining on ImageNet  \citep{deng2009imagenet}. 
The two hyperparameters introduced in equ.~\ref{equ:0} are set to $\lambda_{pos} = 1.0,$ and $\lambda_{seg} = 3.0$, and we use $\omega = 3.0 $ for the weight associated with the influence of depth equ.~\ref{equ:4}. 
These values were determined in preliminary experiments. 
We apply mean fusion to combine RGB and depth features (cf.\ Section~\ref{subsec:colordepthfusion}) and  use  a feature dimension of $C_e$ = 256 for the resultant feature maps (cf.\ Section~\ref{sec:panoptic_fcn}); this is larger than $C_e$ = 64, the value used in \citep{li2021fully}. 
In the following, we refer to the variant of our methodology   trained and parameterised as described in this section as \textit{Ours}. 
All experiments are carried out on a Nvidia A100 GPU with $40$ GB memory. 


\subsubsection{Evaluation Protocol: } \label{subsubsec:evaluationprotocol}

We follow the evaluation scheme of \citet{kirillov2019panoptic}, using the \textit{panoptic quality} ($PQ$) as a quality measure:
\begin{equation}\label{equ:PQ}
    \begin{split}
        PQ= \frac{\sum_{(Pr,Gt)\in TP} IoU\left(Pr,Gt\right)}{|TP|+\frac{1}{2}\cdot|FP|+\frac{1}{2}\cdot|FN|}\,,
    \end{split}
\end{equation}
where $Pr$ and $Gt$ are a predicted and a ground truth mask found to correspond to each other and $IoU$ denotes the Intersection over Union of these masks. 
$TP$ indicates the set of true positive masks, i.e. the set of  masks  $Pr$ for which a ground truth mask with an $IoU >50\%$ could be found. 
Similarly, $FP$ and $FN$ denote the set of false positive masks (e.g.\ predicted {\em thing} instances without a match in the ground truth) and false negative masks (e.g.\ ground truth {\em thing} instances without correspondence in the predictions). 
In addition to $PQ$ we also report the  panoptic quality obtained only for \textit{thing} ($PQ^{th}$) and \textit{stuff} classes ($PQ^{st}$).
Details about the way in which $PQ$, $PQ^{th}$ and $PQ^{st}$ are  determined can be found in  \citep{kirillov2019panoptic}. 


\subsection{Results and Discussion} \label{subsec:evaldepthawarediceloss}

\begin{figure*}[ht]
\begin{center}
		\includegraphics[width=\textwidth]{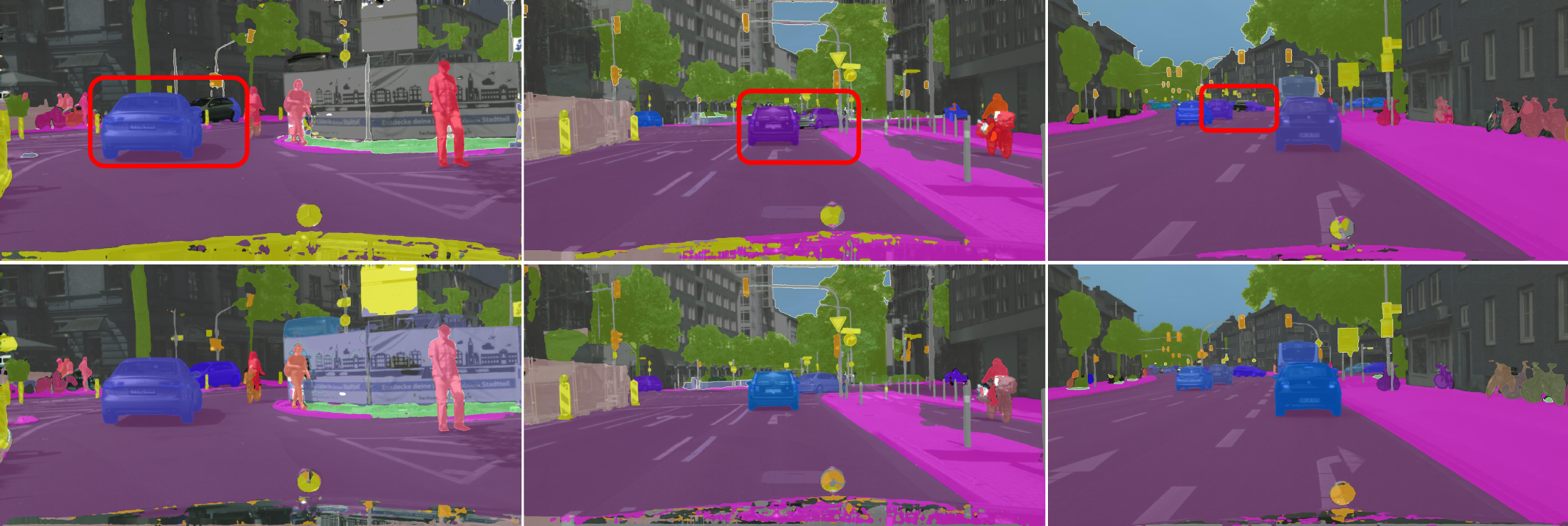}
	\caption{Qualitative examples of results achieved on the Cityscapes data. Top: results of the Panoptic FCN  baseline \protect\citep{li2021fully}, bottom: results of our method. Different colours are used to identify {\em stuff} class or {\em thing} instance to which a pixel is assigned, and the resultant label maps are superimposed to the RGB input images. The red boxes highlight examples in which the baseline erroneously merged two {\em thing} instances, whereas our method separated them correctly. }
\label{fig:compare}
\end{center}
\end{figure*}

Tab.~\ref{tab:tab4} shows the quality metrics achieved by our method ({\em Ours}) on the Cityscapes validation set, and Fig.~\ref{fig:compare} shows some qualitative examples. 
The table also presents the results for two  baseline methods,  \citep{li2021fully} and \citep{de2023intra}. 
We chose \citep{li2021fully}  for comparison because our method is an extension of that method, so that the comparison will highlight the impact of our modifications. 
We trained that baseline using the protocol described in Section~\ref{subsubsec:experimentalsetup}, i.e.\ using $C_e=256$.  
Fig.~\ref{fig:compare} also shows  some qualitative results produced by this baseline method. 
The second baseline \citep{de2023intra} was chosen because it tries to solve the same problem of \citep{li2021fully}  as our method, but using a different strategy (and also not using depth). 
In this case, the quality indices are those published in \citep{de2023intra}, which are based on the same definition of training and test images as ours. 

\begin{table}[ht]
	\centering
		\begin{tabular}{|l|c|c|c|c|}\hline
			Method&$PQ$&$PQ^{th}$&$PQ^{st}$\\\hline
{\em\citep{li2021fully}}&$60.4$&$53.6$&$65.4$ \\ 
             {\em \citep{de2023intra}}&$60.8$&$54.7$&$65.3$ \\ 
             {\em Ours}&$\mathbf{62.6}$&$\mathbf{56.2}$&$\mathbf{67.3}$ \\ \hline
		\end{tabular}
	\caption{Panoptic Quality for all classes ($PQ$) and  for {\em thing}  ($PQ^{th}$) and {\em stuff} ($PQ^{st}$) classes achieved by our method and two baselines. All values are given in $[\%]$. }
\label{tab:tab4}
\end{table}

Note that the values for  \citep{li2021fully} are better than those published in the original paper, probably due to the use of another value for the feature dimension $C_e$ (the minibatch size and number of training iterations we used were also different). 
Compared  to \citep{li2021fully}, the results of  \citep{de2023intra} are slightly better  for $PQ^{th}$ and $PQ$, but slightly lower for $PQ^{st}$. 
Our method outperforms both methods in all quality indices, i.e. both for {\em thing} and {\em stuff} classes. 
Compared to \citep{li2021fully}, the improvement for  {\em thing} classes is more pronounced (+2.6\%) than the one for {\em stuff} classes (+1.9\%), yielding a total improvement in $PQ$ of +2.2\%. 
Compared to \citep{de2023intra}, where the problem of merged instances is tackled as well, the improvement for  {\em thing} classes is still +1.5\%. 
In total, the gain in $PQ$ of our method is +1.8\%. 
We believe that these numbers confirm the hypothesis made in the beginning, namely that the consideration of depth supports the differentiation of {\em thing} instances of similar appearance and, thus, improves the quality of panoptic segmentation. 
This positive effect can also be seen in the areas highlighted by red boxes in Fig.~\ref{fig:compare}. 
Whereas  \citep{li2021fully} tends to assign pixels located on visually similar but distinct {\em thing} instances at different depth levels to the same instance mask, our method mitigates this effect and is able to differentiate such instances.

\begin{figure}[ht]
\begin{center}
		\includegraphics[width=1.0\columnwidth]{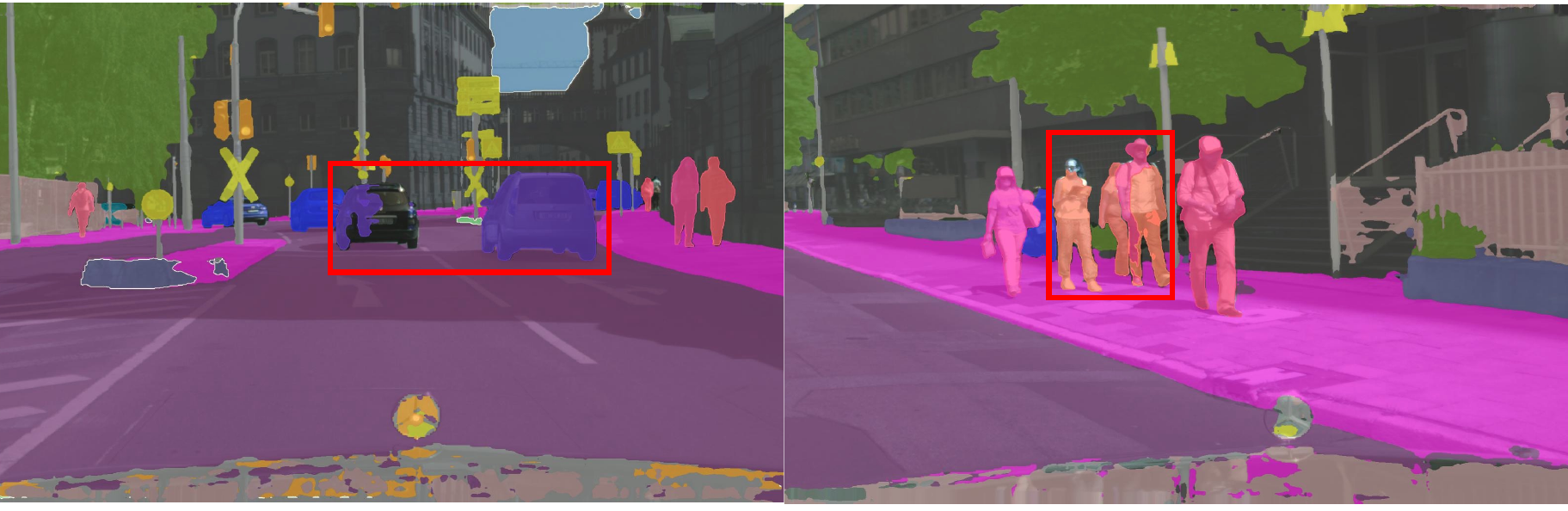}
	\caption{Failure cases of our method: the red boxes indicate instances erroneously merged by our method. The merged instances occur at a similar depth.}
\label{fig:failcases}
\end{center}
\end{figure}

However, there are also some remaining problems.  
In our depth-aware Dice Loss function, the difference in the distances between camera and objects is used to identify distinct {\em thing} instances.
As a result, instances looking similar and occurring at similar distances remain problematic, as shown in Fig.~\ref{fig:failcases}. 
In this case, the depth information does not lead to a further penalisation of FP instance pixels in the loss function compared to the plain dice loss, leading to problems that are similar to those of the baseline \citep{li2021fully}. 
We aim to address this problem in future work, e.g., by including a penalty based on the 3D distance between distinct instances in the loss function  instead of only relying on the difference in depth.

\subsection{Ablation Studies} \label{subsec:Ablation}


\subsubsection{Influence of the  weight $\omega$: } \label{subsec:omegaablation}
In this section, we investigate the influence of the weight $\omega$  associated with the influence of depth information in our loss function (cf.~Sec.\,\ref{subsec:depthawarediceloss}) on the performance of our method.
For this purpose, we trained our method several times in the way described in Section~\ref{subsubsec:experimentalsetup}, using  different values for $\omega$, namely $0, 1, 3, 5$ and $10$. 
Note that $\omega=3$ is the setting analysed in the previous section.
In the setting with $\omega=0$, the original Dice loss is used for training, i.e., depth is used as an additional input, but training is based on the loss used in \citep{li2021fully}. 
The results are shown in Tab.~\ref{tab:tab5}. 
In general, the differences in $PQ$ are in the order of 1\%. 
The influence of $\omega$ is larger for {\em thing} classes than it is for {\em stuff}. 
Using $\omega=3$ achieves the best results with respect all compared quality metrics. 
It is particularly interesting to compare this result to the one achieved when the original Dice loss is used ($\omega=0$). 
The quality indices in Tab.~\ref{tab:tab5} show that using the extension of the Dice Loss leads to an improvement of $PQ$ independently from the value of $\omega$ used, and in case of $\omega=3$, the improvement is +1.5\%. 
Interestingly, the $PQ$ values for both, the {\em thing} and {\em stuff} classes are positively affected, even though the depth-aware Dice loss is only applied to  {\em thing} instances in training; probably a smaller number of FP instances leads to a lower error rate for {\em stuff} pixels, which would affect the $PQ^{st}$ metric via the $IoU$ values in equ.~\ref{equ:PQ}. 
Nevertheless, the improvement in $PQ^{th}$ (+2.7\%) is larger than the one in  $PQ^{st}$ (+1.2\%), probably for that very reason. 
On the other hand, the $PQ$ metric achieved using $\omega=0$ in Tab.~\ref{tab:tab5} is still slightly better than the one reported for both baselines in Tab.~\ref{tab:tab4}, which is largely due to an improvement of the segmentation quality for {\em stuff} classes, as indicated by the $PQ^{st}$ values. 
We can conclude that just using depth as an additional input improves the results slightly, mainly for {\em stuff} classes; introducing the depth-aware Dice loss in training further improves the results, in this case with a larger impact on the {\em thing} classes. 

\begin{table}[ht]
	\centering
		\begin{tabular}{|l|c|c|c|c|}\hline
			$\omega$&$PQ$&$PQ^{th}$&$PQ^{st}$\\\hline
                $0$&$61.1$&$53.5$&$66.1$ \\
                $1$&$61.5$&$53.3$&$67.5$ \\
                $3$&$62.6$&$56.2$&$67.3$ \\ 
                $5$&$62.1$&$55.0$&$67.2$ \\ 
                $10$&$61.6$&$54.8$&$66.6$ \\ \hline
		\end{tabular}
	\caption{Quality metrics $[\%]$ achieved when training our method with different values of the hyperparameter $\omega$ in equ.~\ref{equ:4}.}
\label{tab:tab5}
\end{table}


\subsubsection{Comparison of fusion schemes: } \label{subsec:evaldepthcolorfusion}
In this section, we investigate the influence of the fusion scheme used for combining features extracted from the RGB images and the depth maps (cf.~Section~\ref{subsec:colordepthfusion}). 
For this purpose, we compare the results achieved using the mean fusion scheme (equ.\,\ref{equ:1_1}), which were already discussed  in Section~\ref{subsec:evaldepthawarediceloss}, to those achieved when applying fusion based on  concatenation. 
In order to obtain the latter, another model was trained, using the protocol described in Section~\ref{subsubsec:experimentalsetup} but replacing mean fusion by the fusion scheme according to equ.~\ref{equ:1_2}. 
The results are shown in Tab.~\ref{tab:tab3}.
The quality indices in  Tab.~\ref{tab:tab3} indicate that mean fusion is to be preferred: the $PQ$ is better by 1.1\% when using mean fusion, and the other indices are also higher for that variant. For {\em thing} instances ($PQ^{th}$), the difference is $2.1\%$. 

\begin{table}[ht]
	\centering
		\begin{tabular}{|l|c|c|c|}\hline
			Fusion&$PQ$&$PQ^{th}$&$PQ^{st}$\\\hline
             \textit{mean}&$62.6$&$56.2$&$67.3$ \\ 
             \textit{concatenation}&$61.5$&$54.1$&$67.0$ \\ \hline
		\end{tabular}
	\caption{Quality metrics $[\%]$ achieved when using different fusion schemes for combining RGB and depth features: \textit{mean} fusion (equ.~\ref{equ:1_1}) and fusion by \textit{concatenation}  (equ.~\ref{equ:1_2}).
  The values for mean fusion are identical to those in Tab.~\ref{tab:tab4}.}
\label{tab:tab3}
\end{table}


\section{Conclusion}\label{sec:CONCLUSION}

In this paper, we  present a new CNN-based method for panoptic segmentation which combines colour and depth information to overcome problems of existing methods  based on RGB images only. 
Depth is considered in two ways. 
On the one hand, depth is processed along with RGB images in separate network branches, and the resultant feature maps are combined  in a late fusion approach. 
On the other hand, our method is based on a new depth-aware dice loss term which penalises the assignment of pixels to the same \textit{thing} instance based on the difference between their associated depth values.
Experiments carried out on the Cityscapes dataset show that the proposed method outperforms the baseline method in terms of panoptic quality by $+2.2\%$ in total and by $+2.6\%$ and $+1.9\%$ for \textit{thing} and \textit{stuff} classes, respectively.
The improvements in \textit{thing} classes are mainly achieved by a reduction in the number of objects that are erroneously merged into one \textit{thing} instance.
Our results confirm that it is beneficial to   consider explicit 3D information about the scene in  panoptic segmentation.

As we use the difference depth to compute a penalty term in our loss function, the correct segmentation of distinct objects of similar appearance located at the same depth remain a challenge.
We want to  address this problem in future work by including a penalty term into the loss function that based on the 3D distance between distinct objects.
Moreover, we plan to extend the presented method by incorporating temporal information, i.e., by using sequences of  images with associated depth maps instead of data acquired at a single point in time.



\section*{Acknowledgements}\label{sec:ACKNOWLEDGEMENTS}
This work was supported by the German Research Foundation (DFG) as a part of the Research Training Group i.c.sens [GRK2159]. 
Computations were carried out on the LUH computer cluster, funded by the Leibniz Universität Hannover, the Lower Saxony Ministry of Science and Culture (MWK), and  DFG.

{
	\begin{spacing}{1.17}
	   \normalsize
	   \bibliography{ISPRSguidelines_authors} 

\begin{thebibliography}{xx}

\bibitem[Cheng et al., 2020]{cheng2020panoptic}
Cheng, B., Collins, M.~D., Zhu, Y., Liu, T., Huang, T.~S., Adam, H., Chen, L.-C., 2020.
 Panoptic-deeplab: A simple, strong, and fast baseline for bottom-up panoptic segmentation.
 \emph{Proceedings of the IEEE Conference on Computer Vision and Pattern Recognition (CVPR)}, 12475--12485.

\bibitem[Cordts et al., 2016]{cordts2016cityscapes}
Cordts, M., Omran, M., Ramos, S., Rehfeld, T., Enzweiler, M., Benenson, R., Franke, U., Roth, S., Schiele, B., 2016.
 The cityscapes dataset for semantic urban scene understanding.
 \emph{Proceedings of the IEEE Conference on Computer Vision and Pattern Recognition (CVPR)}, 3213--3223.

\bibitem[de~Geus and Dubbelman, 2023]{de2023intra}
de~Geus, D., Dubbelman, G., 2023.
 Intra-batch supervision for panoptic segmentation on high-resolution images.
 \emph{Proceedings of the IEEE/CVF Winter Conference on Applications of Computer Vision}, 3165--3173.

\bibitem[Deng et al., 2009]{deng2009imagenet}
Deng, J., Dong, W., Socher, R., Li, L.-J., Li, K., Fei-Fei, L., 2009.
 Imagenet: A large-scale hierarchical image database.
 \emph{Proceedings of the IEEE Conference on Computer Vision and Pattern Recognition (CVPR)}, 248--255.

\bibitem[Hirschmuller, 2007]{hirschmuller2007stereo}
Hirschmuller, H., 2007.
 Stereo processing by semiglobal matching and mutual information.
 {\em IEEE Transactions on Pattern Analysis and Machine Intelligence}, 30(2), 328--341.

\bibitem[Hu et al., 2018]{hu2018squeeze}
Hu, J., Shen, L., Sun, G., 2018.
 Squeeze-and-excitation networks.
 \emph{Proceedings of the IEEE Conference on Computer Vision and Pattern Recognition (CVPR)}, 7132--7141.

\bibitem[Kirillov et al., 2019a]{kirillov2019panopticfpn}
Kirillov, A., Girshick, R., He, K., Doll{\'a}r, P., 2019a.
 Panoptic feature pyramid networks.
 \emph{Proceedings of the IEEE Conference on Computer Vision and Pattern Recognition (CVPR)}, 6399--6408.

\bibitem[Kirillov et al., 2019b]{kirillov2019panoptic}
Kirillov, A., He, K., Girshick, R., Rother, C., Doll{\'a}r, P., 2019b.
 Panoptic segmentation.
 \emph{Proceedings of the IEEE Conference on Computer Vision and Pattern Recognition (CVPR)}, 9404--9413.

\bibitem[Li et al., 2019]{li2019attention}
Li, Y., Chen, X., Zhu, Z., Xie, L., Huang, G., Du, D., Wang, X., 2019.
 Attention-guided unified network for panoptic segmentation.
 \emph{Proceedings of the IEEE Conference on Computer Vision and Pattern Recognition (CVPR)}, 7026--7035.

\bibitem[Li et al., 2021]{li2021fully}
Li, Y., Zhao, H., Qi, X., Wang, L., Li, Z., Sun, J., Jia, J., 2021.
 Fully convolutional networks for panoptic segmentation.
 \emph{Proceedings of the IEEE Conference on Computer Vision and Pattern Recognition (CVPR)}, 214--223.

\bibitem[Lin et al., 2017a]{lin2017feature}
Lin, T.-Y., Doll{\'a}r, P., Girshick, R., He, K., Hariharan, B., Belongie, S., 2017a.
 Feature pyramid networks for object detection.
 \emph{Proceedings of the IEEE Conference on Computer Vision and Pattern Recognition (CVPR)}, 2117--2125.

\bibitem[Lin et al., 2017b]{lin2017focal}
Lin, T.-Y., Goyal, P., Girshick, R., He, K., Doll{\'a}r, P., 2017b.
 Focal loss for dense object detection.
 \emph{Proceedings of the IEEE Conference on Computer Vision and Pattern Recognition (CVPR)}, 2980--2988.

\bibitem[Lipson et al., 2021]{lipson2021raft}
Lipson, L., Teed, Z., Deng, J., 2021.
 Raft-stereo: Multilevel recurrent field transforms for stereo matching.
 \emph{Proceedings of the International Conference on 3D Vision (3DV)}, 218--227.

\bibitem[Milletari et al., 2016]{milletari2016v}
Milletari, F., Navab, N., Ahmadi, S.-A., 2016.
 V-net: Fully convolutional neural networks for volumetric medical image segmentation.
 \emph{Proceedings of the International Conference on 3D Vision (3DV)}, 565--571.

\bibitem[Narita et al., 2019]{narita2019panopticfusion}
Narita, G., Seno, T., Ishikawa, T., Kaji, Y., 2019.
 Panopticfusion: Online volumetric semantic mapping at the level of stuff and things.
 \emph{IEEE/RSJ International Conference on Intelligent Robots and Systems (IROS)}, 4205--4212.

\bibitem[Seichter et al., 2022]{seichter2022efficient}
Seichter, D., Fischedick, S.~B., K{\"o}hler, M., Gro{\ss}, H.-M., 2022.
 Efficient multi-task rgb-d scene analysis for indoor environments.
 \emph{International Joint Conference on Neural Networks (IJCNN)}, 1--10.

\bibitem[Seichter et al., 2021]{seichter2021efficient}
Seichter, D., K{\"o}hler, M., Lewandowski, B., Wengefeld, T., Gross, H.-M., 2021.
 Efficient {RGB-D} semantic segmentation for indoor scene analysis.
 \emph{2021 IEEE International Conference on Robotics and Automation (ICRA)}, 13525--13531.

\bibitem[Wang et al., 2021]{wang2021max}
Wang, H., Zhu, Y., Adam, H., Yuille, A., Chen, L.-C., 2021.
 Max-deeplab: End-to-end panoptic segmentation with mask transformers.
 \emph{Proceedings of the IEEE Conference on Computer Vision and Pattern Recognition (CVPR)}, 5463--5474.

\bibitem[Wu et al., 2021]{wu2021scenegraphfusion}
Wu, S.-C., Wald, J., Tateno, K., Navab, N., Tombari, F., 2021.
 Scenegraphfusion: Incremental 3d scene graph prediction from rgb-d sequences.
 \emph{Proceedings of the IEEE Conference on Computer Vision and Pattern Recognition (CVPR)}, 7515--7525.

\bibitem[Xiong et al., 2019]{xiong2019upsnet}
Xiong, Y., Liao, R., Zhao, H., Hu, R., Bai, M., Yumer, E., Urtasun, R., 2019.
 {UPS}net: A unified panoptic segmentation network.
 \emph{Proceedings of the IEEE Conference on Computer Vision and Pattern Recognition (CVPR)}, 8818--8826.

\bibitem[Yu et al., 2022]{yu2022cmt}
Yu, Q., Wang, H., Kim, D., Qiao, S., Collins, M., Zhu, Y., Adam, H., Yuille, A., Chen, L.-C., 2022.
 Cmt-deeplab: Clustering mask transformers for panoptic segmentation.
 \emph{Proceedings of the IEEE Conference on Computer Vision and Pattern Recognition (CVPR)}, 2560--2570.

\bibitem[Zhang et al., 2021]{zhang2021k}
Zhang, W., Pang, J., Chen, K., Loy, C.~C., 2021.
 K-net: Towards unified image segmentation.
 {\em Advances in Neural Information Processing Systems}, 34, 10326--10338.

\bibitem[Zhou et al., 2019]{zhou2019objects}
Zhou, X., Wang, D., Kr{\"a}henb{\"u}hl, P., 2019.
 Objects as points.
 {\em arXiv preprint arXiv:1904.07850}.

\end{thebibliography}
	\end{spacing}
}

\end{document}